\documentclass[conference]{IEEEtran}
\IEEEoverridecommandlockouts
\usepackage{cite}
\usepackage{amsmath,amssymb,amsfonts}
\usepackage{wrapfig}
\usepackage{algorithmic}
\usepackage{algorithm}
\usepackage{tabularx} 
\usepackage{booktabs}
\usepackage{graphicx}
\usepackage{textcomp}
\usepackage{xcolor}
\usepackage{booktabs}
\usepackage{multirow}

\def\BibTeX{{\rm B\kern-.05em{\sc i\kern-.025em b}\kern-.08em
    T\kern-.1667em\lower.7ex\hbox{E}\kern-.125emX}}
\begin{document}

\title{SARD: Segmentation-Aware Anomaly Synthesis via Region-Constrained Diffusion with Discriminative Mask Guidance}

% \author{
%     ${\textnormal{Yanshu Wang}}^{1}$,
%     ${\textnormal{Xichen Xu}}^{1}$,
%     ${\textnormal{Xiaoning Lei}}^{2}$,
%     and ${\textnormal{Guoyang Xie}}^{2}$\\
%     $^1$\textit{Global Institute of Future Technology, Shanghai Jiao Tong University}, Shanghai, China \\
%     $^2$\textit{Department of Intelligent Manufacturing,  Contemporary Amperex Technology Co.,Ltd}, Ningde, China \\
%     isaac\_wang@sjtu.edu.cn, neptune\_2333@sjtu.edu.cn, leixn01@outlook.com, guoyang.xie@ieee.org
% }

\author{
\IEEEauthorblockN{Yanshu Wang}
\IEEEauthorblockA{\textit{Global Institute of Future Technology}\\
\textit{Shanghai Jiao Tong University}\\
Shanghai, China\\
\texttt{isaac\_wang@sjtu.edu.cn}}
\\
\IEEEauthorblockN{Xiaoning Lei}
\IEEEauthorblockA{\textit{Department of Intelligent Manufacturing}\\
\textit{Contemporary Amperex Technology Co.,Ltd}\\
Ningde, China\\
\texttt{leixn01@outlook.com}}
\and
\IEEEauthorblockN{Xichen Xu}
\IEEEauthorblockA{\textit{Global Institute of Future Technology}\\
\textit{Shanghai Jiao Tong University}\\
Shanghai, China\\
\texttt{neptune\_2333@sjtu.edu.cn}}
\\
\IEEEauthorblockN{Guoyang Xie}
\IEEEauthorblockA{\textit{Department of Intelligent Manufacturing}\\
\textit{Contemporary Amperex Technology Co.,Ltd}\\
Ningde, China\\
\texttt{guoyang.xie@ieee.org}}
}

\vspace{-42pt}

\maketitle

\begin{abstract}
Synthesizing anomalies that are both realistic and \emph{spatially precise} is most valuable when it directly benefits pixel-wise segmentation. Recent diffusion-based methods show strong generative capacity but often lack spatial controllability and can unintentionally alter background content. We propose \textbf{SARD} (Segmentation-Aware Anomaly Synthesis via Region-Constrained Diffusion with Discriminative Mask Guidance), a diffusion-based framework tailored for segmentation-oriented anomaly generation. Specifically, \emph{Region-Constrained Diffusion} (RCD) performs a \emph{masked reverse update}: the posterior sample is written only inside the anomaly mask while the background is \emph{passed through} from the current noisy input, preventing background drift outside the target region. In addition, a \emph{Discriminative Mask Guidance} (DMG) module augments the discriminator with a mask-aware branch that supplies localized adversarial feedback for sharper textures and boundary alignment. Extensive experiments on MVTec-AD and BTAD, across SegFormer and BiSeNet~V2 backbones, show consistent gains in mIoU and pixel accuracy over strong baselines. By explicitly coupling background preservation with mask-guided discrimination, SARD advances \emph{segmentation-aware anomaly synthesis}.
\end{abstract}


\section{Introduction}
\begin{figure*}[ht]
  \centering
  \includegraphics[width=0.96\textwidth]{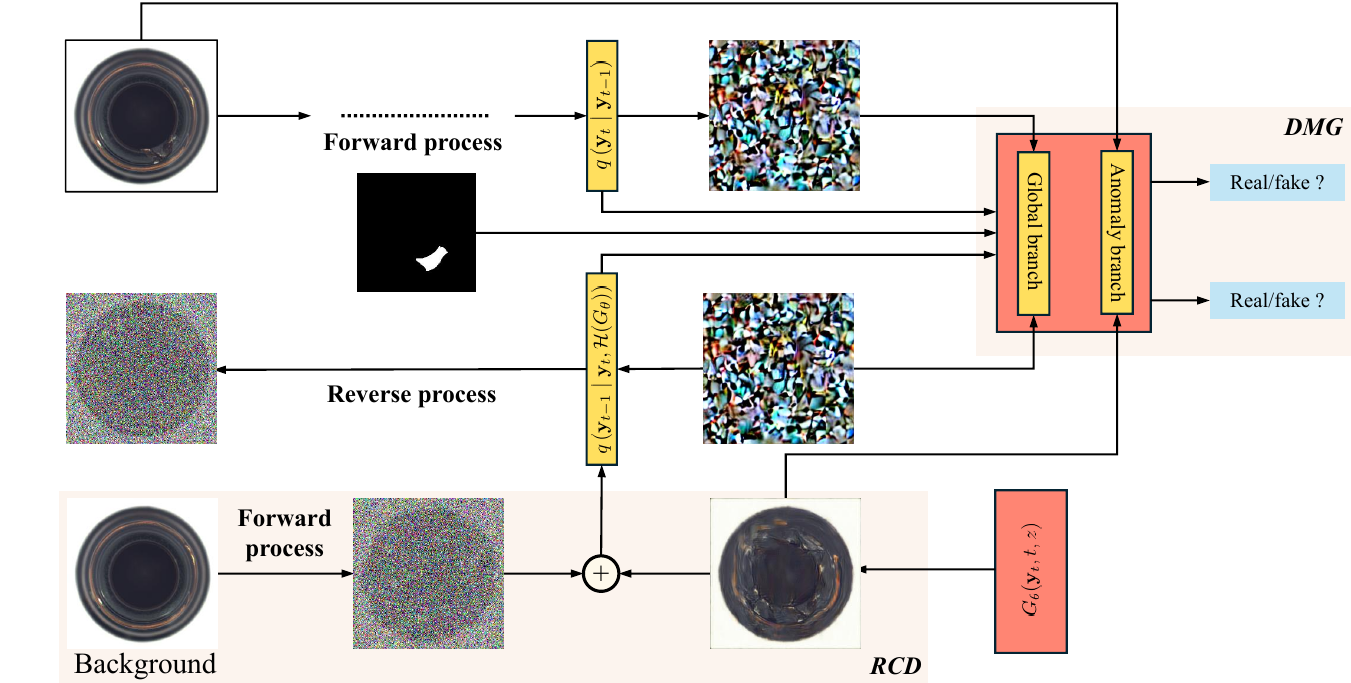}
    \caption{
    Overview of the proposed SARD framework. At each reverse diffusion step, the generator predicts a clean image, which is used to sample a posterior estimate. Region-Constrained Diffusion (RCD) fuses this sample with the noisy input, preserving background regions while updating only the anomaly area. A dual-branch discriminator enforces both global realism and mask-guided local fidelity.
    }

  \label{fig:framework}
\end{figure*}

Industrial anomaly segmentation is essential for automated visual inspection in modern manufacturing, as it enables fine-grained pixel-level localization of defects. Compared to image-level detection, segmentation offers significantly more actionable insights for high-resolution quality control. However, acquiring large-scale, accurately annotated segmentation masks is notoriously challenging due to the rarity, diversity, and annotation costs associated with real-world anomalies. This limitation has spurred increasing interest in synthetic anomaly generation to augment training data and enhance segmentation performance.

Prior works on anomaly synthesis typically rely on rule-based manipulations such as patch replacement or texture blending~\cite{li2021cutpaste, zavrtanik2021draem}. While simple and interpretable, these approaches provide limited control over where and how defects are placed, and they often struggle to preserve structural or semantic consistency at object boundaries, especially under large appearance or material changes. GAN-based methods~\cite{niu2020defect, dfmgan} improve generation fidelity but still suffer from unstable training, weak spatial controllability, and difficulty aligning synthesized defects with specified mask regions—particularly for small, localized anomalies.

Recently, denoising diffusion models~\cite{song2021scorebased, rombach2022high} have demonstrated superior generative performance with stable optimization and high sample quality. However, most diffusion-based anomaly generators operate on the entire image without explicitly decoupling foreground defects from background content. This whole-image treatment can cause background drift and make the synthesized anomalies blurry or misaligned with the given masks, resulting in a spatial mismatch between defects and their designated locations.

To overcome these challenges, we propose \textbf{SARD} (Segmentation-Aware Anomaly Synthesis via Region-Constrained Diffusion), a novel diffusion-based framework tailored for pixel-level anomaly synthesis. Our approach introduces two key innovations:

\begin{itemize}
\item \textbf{Region-Constrained Diffusion (RCD):} Instead of updating the entire image during reverse diffusion, RCD performs a masked reverse update: the posterior sample is written only inside the anomaly mask, while the background is directly passed through from the current noisy input. This prevents unintended background changes and focuses the generative capacity on learning defect textures within the target region.

\item \textbf{Discriminative Mask Guidance (DMG):} We design a dual-branch discriminator that jointly evaluates global realism and region-specific fidelity. The foreground-aware branch leverages binary anomaly masks to compute localized adversarial losses, guiding the generator to produce sharper and mask-aligned anomalies.
\end{itemize}

We evaluate \textbf{SARD} on two benchmark datasets: MVTec-AD~\cite{bergmann2019mvtec} and BTAD~\cite{btad2020}, using SegFormer~\cite{xie2021segformer} and BiseNet V2~\cite{yu2021bisenet} as real-time segmentation backbones. Extensive comparisons with representative methods—including CutPaste~\cite{li2021cutpaste}, DRAEM~\cite{zavrtanik2021draem}, GLASS~\cite{glass}, DFMGAN~\cite{dfmgan}, RealNet~\cite{zhang2024realnet}, AnomalyDiffusion~\cite{anomalydiffusion}, and DDGAN~\cite{ddgan2022}, show that SARD significantly improves downstream segmentation performance.

Furthermore, ablation studies confirm the effectiveness of RCD and DMG in enhancing spatial alignment, texture realism, and segmentation-friendliness of the generated anomalies. By explicitly modeling foreground-background separation and introducing region-specific supervision, SARD establishes a new state-of-the-art in diffusion-based anomaly synthesis for industrial visual inspection.

% 对应的这里改一哈
\section{Related Work}
\textbf{Anomaly Synthesis.}
To alleviate the scarcity of annotated industrial anomalies, a variety of strategies have been explored for synthetic anomaly generation in recent years. Early training-free approaches, such as CutPaste and DRAEM~\cite{li2021cutpaste, zavrtanik2021draem}, rely on rule-based manipulations like patch replacement or image reconstruction to simulate anomaly patterns. While simple and interpretable, these approaches provide limited control over where and how defects are placed and can weaken structural or semantic consistency at object boundaries under large appearance or material changes.

GAN-based methods~\cite{niu2020defect, du2022new}, including representative approaches such as DFMGAN~\cite{dfmgan}, enhance generation fidelity through adversarial training, yet they can suffer from unstable optimization, limited spatial controllability, and difficulties in aligning synthesized defects with designated mask regions—particularly for small, localized anomalies.

Diffusion-based methods~\cite{ho2020denoising, rombach2022high} have recently emerged as strong alternatives, offering stable training dynamics and high-fidelity synthesis through iterative refinement. For example, DDGAN~\cite{ddgan2022} introduces adversarial feedback into the reverse process to accelerate sampling, whereas AnomalyDiffusion~\cite{anomalydiffusion} leverages diffusion priors for anomaly representation. Nonetheless, many approaches still operate on the whole image without explicitly decoupling foreground defects from background content, which can cause background drift and spatial mismatch between synthesized defects and target masks. Segmentation-aware techniques such as GLASS~\cite{glass} and Synth4Seg~\cite{mou2024synth4seg} incorporate geometric constraints or mask supervision, yet the extent of foreground–background decoupling during reverse updates remains limited in practice.

\textbf{Anomaly Segmentation.}
Pixel-wise anomaly segmentation plays a critical role in industrial inspection and visual quality control, yet remains difficult due to the rarity, diversity, and subtle appearance of real-world defects. Traditional approaches, including both reconstruction-based and feature-based pipelines~\cite{xu2023fascinating, yang2024slsg, liu2024dual}, often rely on assumptions of normality but struggle to accurately localize small-scale, sparse, or low-contrast anomalies. These limitations are particularly pronounced in fine-grained defect scenarios.

To overcome these challenges, synthetic data augmentation has emerged as a promising direction. Methods such as GLASS, Synth4Seg, and RealNet utilize synthesized anomalies during training to improve segmentation performance under limited annotation conditions. These works emphasize the importance of structural realism, spatial precision, and mask fidelity in the generation process. Evaluations are typically carried out using lightweight and real-time segmentation backbones, including SegFormer~\cite{xie2021segformer} and BiseNet V2~\cite{yu2021bisenet}, ensuring computational efficiency and consistent comparisons across benchmarks.

\section{Proposed Method}

Given normal images and their corresponding anomaly masks, our goal is to generate synthetic anomalies that are both realistic and precisely aligned with the masks, enabling downstream segmentation models to learn robust detection boundaries. To this end, we introduce a novel diffusion-based framework comprising two key components: (\romannumeral1) \textbf{Region-Constrained Diffusion (RCD)} — a modified reverse sampling mechanism that preserves the real background and restricts generation to the foreground anomaly regions; and (\romannumeral2) \textbf{Discriminative Mask Guidance (DMG)} — a discriminator structure that introduces region-specific supervision by focusing on both the entire image and the masked foreground anomaly area. An overview of the framework is illustrated in Fig.~\ref{fig:framework}.

\subsection{Region-Constrained Diffusion (RCD)}
In denoising diffusion probabilistic models (DDPM)~\cite{ho2020denoising}, each reverse step aims to sample a cleaner image \( \mathbf{x}_{t-1} \) from a noisy input \( \mathbf{x}_t \). The model defines the posterior distribution over \( \mathbf{x}_{t-1} \) as a Gaussian conditioned on the estimated clean image \( \hat{\mathbf{x}}_0 \):
\begin{align}
\boldsymbol{\mu}_t &= A_t \hat{\mathbf{x}}_0 + B_t \mathbf{x}_t,  \\
\boldsymbol{\Sigma}_t &= \sigma_t^2 \mathbf{I},  \\
q(\mathbf{x}_{t-1} \mid \mathbf{x}_t, \hat{\mathbf{x}}_0) &= \mathcal{N}(\boldsymbol{\mu}_t, \boldsymbol{\Sigma}_t), 
\end{align}
where \( A_t \), \( B_t \), and \( \sigma_t^2 \) are timestep-dependent constants derived from the forward diffusion process. These expressions allow for analytical sampling as long as a reliable prediction of \( \hat{\mathbf{x}}_0 \) is available.

While DDPM theoretically assumes a Gaussian forward process to enable these closed-form expressions, the assumption becomes less accurate at lower timesteps or in complex data regimes. To flexibly approximate the unknown clean image, we employ a learnable generator \( G \) that directly predicts \( \hat{\mathbf{x}}_0 \) from noisy input:
\begin{equation}
\hat{\mathbf{x}}_0 = G(\mathbf{x}_t, t, \mathbf{z}),
\end{equation}
where \( \mathbf{z} \sim \mathcal{N}(0, \mathbf{I}) \) is a latent variable that introduces sample diversity.

However, this standard sampling strategy indiscriminately updates all spatial regions of the image, which is suboptimal for anomaly synthesis. In industrial scenarios, anomalies are typically sparse and localized, and modifying the entire image may degrade background fidelity.

To address this, we introduce \textbf{Region-Constrained Diffusion (RCD)}, which incorporates a binary anomaly mask \( \mathbf{m} \in \{0,1\}^{H \times W} \) to restrict updates to anomaly regions. The denoised result is fused with the noisy input as:
\begin{equation}
\mathbf{x}_{t-1} = \mathbf{m} \odot \mathbf{x}_{t-1}' + (1 - \mathbf{m}) \odot \mathbf{x}_t, 
\end{equation}
where \( \odot \) denotes element-wise multiplication. This operation preserves the forward trajectory in background regions while allowing targeted refinement of the anomaly areas.

% We summarize the masked update as:
% %
% \begin{equation}
% \mathbf{x}_{t-1} = \mathcal{T}_\mathrm{rcd}(\mathbf{x}_t, \hat{\mathbf{x}}_0, \mathbf{m}) := \mathbf{m} \odot \mathbf{x}_{t-1}' + (1 - \mathbf{m}) \odot \mathbf{x}_t. \tag{7}
% \end{equation}

Because the mask is applied post-sampling, the underlying DDPM formulation remains intact. This simple yet effective mechanism enables spatially controlled generation of anomalies without compromising the realism and consistency of the background.

\subsection{Discriminative Mask Guidance (DMG)}

Effective adversarial training relies on the discriminator to provide precise and informative feedback to the generator. In conventional GANs, the discriminator typically operates at the image level, focusing on holistic realism. However, such global supervision often fails to capture subtle, spatially sparse anomalies—particularly in industrial scenarios where defects are small and localized. This motivates the need for a region-sensitive feedback mechanism that can explicitly target foreground anomaly areas.

To this end, we propose \textbf{Discriminative Mask Guidance (DMG)}, a dual-branch discriminator structure that jointly evaluates image-level realism and region-specific fidelity guided by anomaly masks.

Let the generated image be \( \mathbf{x} \in \mathbb{R}^{C \times H \times W} \), and its associated binary anomaly mask be \( \mathbf{m} \in \{0,1\}^{1 \times H \times W} \). Our discriminator \( D \) produces two parallel outputs:
\begin{equation}
D(\mathbf{x}, \mathbf{m}) = \left(D_{\text{img}}(\mathbf{x}),\; D_{\text{fg}}(\mathbf{x}, \mathbf{m})\right),
\end{equation}
where \( D_{\text{img}}(\mathbf{x}) \) denotes the global discriminator output assessing holistic realism, while \( D_{\text{fg}}(\mathbf{x}, \mathbf{m}) \) evaluates the realism of the synthesized anomaly regions using mask-aware supervision.

To compute \( D_{\text{fg}} \), we extract intermediate activations \( f_l(\mathbf{x}) \in \mathbb{R}^{C' \times H' \times W'} \) from a designated layer \( l \) of the discriminator and apply an upsampled binary mask \( \text{Up}(\mathbf{m}) \in \{0,1\}^{1 \times H' \times W'} \). The mask focuses the auxiliary prediction on foreground anomaly regions:
\begin{equation}
D_{\text{fg}}(\mathbf{x}, \mathbf{m}) = \psi\left(f_l(\mathbf{x}) \odot \text{Up}(\mathbf{m})\right),
\end{equation}
where \( \odot \) denotes element-wise multiplication, and \( \psi(\cdot) \) is a lightweight convolutional head (e.g., one or two layers) applied on masked features. This ensures that discriminator feedback is only activated in regions marked anomalous, promoting targeted supervision.

The overall discriminator loss combines global and foreground components for both real and generated samples:
\begin{align}
\mathcal{L}_D =\; & \mathbb{E}_{(\mathbf{x}, \mathbf{m}) \sim p_r}\left[\phi\big(-D_{\text{img}}(\mathbf{x})\big) + \lambda \cdot \phi\big(-D_{\text{fg}}(\mathbf{x}, \mathbf{m})\big)\right] \nonumber \\
+ & \mathbb{E}_{(\mathbf{x}, \mathbf{m}) \sim p_g}\left[\phi\big(D_{\text{img}}(\mathbf{x})\big) + \lambda \cdot \phi\big(D_{\text{fg}}(\mathbf{x}, \mathbf{m})\big)\right],
\end{align}
where $p_r$ and $p_g$ denote the distributions of real data and generated data, respectively, \( \phi(x) = \log(1 + e^x) \) is the softplus function, and \( \lambda \) is a balancing coefficient (set to 0.2 unless otherwise specified).

During training, both real and synthesized images are paired with binary masks and passed to the dual-branch discriminator. The generator is then optimized to fool both branches by producing samples that are globally realistic and locally consistent with the target anomaly region. This dual perspective effectively enhances anomaly texture quality, boundary precision, and semantic alignment, thereby improving segmentation-aware synthesis performance.

% 下面是更新的
\begin{table*}[htbp]
\centering
\vskip -0.1in
\caption{Comparison of pixel-level anomaly segmentation (mIoU/Acc) using SegFormer trained on synthetic MVTec data produced from the proposed SARD and other existing AS methods.}
\resizebox{1\textwidth}{!}{
\begin{tabular}{l|cc|cc|cc|cc|cc|cc|cc|cc}
\hline
\textbf{Category} 
& \multicolumn{2}{c|}{\textbf{CutPaste}} 
& \multicolumn{2}{c|}{\textbf{DRAEM}} 
& \multicolumn{2}{c|}{\textbf{GLASS}} 
& \multicolumn{2}{c|}{\textbf{RealNet}} 
& \multicolumn{2}{c|}{\textbf{DFMGAN}} 
& \multicolumn{2}{c|}{\textbf{AnomalyDiffusion}} 
& \multicolumn{2}{c|}{\textbf{DDGAN}} 
& \multicolumn{2}{c}{\textbf{SARD}} \\
\hline
& \textbf{mIoU} $\uparrow$ & \textbf{Acc} $\uparrow$ 
& \textbf{mIoU} $\uparrow$ & \textbf{Acc} $\uparrow$ 
& \textbf{mIoU} $\uparrow$ & \textbf{Acc} $\uparrow$ 
& \textbf{mIoU} $\uparrow$ & \textbf{Acc} $\uparrow$ 
& \textbf{mIoU} $\uparrow$ & \textbf{Acc} $\uparrow$ 
& \textbf{mIoU} $\uparrow$ & \textbf{Acc} $\uparrow$ 
& \textbf{mIoU} $\uparrow$ & \textbf{Acc} $\uparrow$ 
& \textbf{mIoU} $\uparrow$ & \textbf{Acc} $\uparrow$ \\
\hline
bottle     & 75.11 & 79.49 & 79.51 & 84.99 & 70.26 & 76.30 & 77.96 & 83.90 & 75.45 & 80.39 & 76.39 & 83.54 & 66.75 & 70.70 & \textbf{83.52} & \textbf{87.71} \\
cable      & 55.40 & 60.49 & 64.52 & 70.77 & 58.81 & 62.32 & 62.51 & 69.27 & 62.10 & 64.87 & 62.49 & 74.48 & 55.83 & 61.67 & \textbf{70.76} & \textbf{76.60} \\
capsule    & 35.15 & 40.29 & 51.39 & 62.32 & 34.12 & 38.04 & 46.76 & 51.91 & 41.29 & 45.83 & 37.73 & 44.72 & 40.42 & 48.74 & \textbf{61.25} & \textbf{71.04} \\
carpet     & 66.34 & 77.59 & 72.57 & 81.28 & 70.11 & 77.56 & 68.84 & 79.15 & 71.33 & 83.69 & 64.67 & 73.59 & 67.98 & 79.25 & \textbf{76.39} & \textbf{84.77} \\
grid       & 29.90 & 46.72 & 47.75 & \textbf{67.85} & 37.43 & 46.30 & 37.55 & 48.86 & 37.73 & 54.13 & 38.70 & 51.82 & 47.56  & 47.99  & \textbf{53.15} & 67.56 \\
hazel\_nut & 56.95 & 60.72 & 84.22 & 89.74 & 55.51 & 57.43 & 60.18 & 63.49 & 83.43 & 86.03 & 59.33 & 67.48 & 63.77 & 68.33 & \textbf{79.30} & \textbf{82.50} \\
leather    & 57.23 & 63.49 & 64.12 & 71.49 & 62.05 & 73.38 & 68.29 & 77.16 & 60.96 & 68.02 & 56.45 & 62.51 & 58.51 & 70.41 & \textbf{72.38} & \textbf{80.16} \\
metal\_nut & 88.78 & 90.94 & 93.51 & 96.10 & 88.15 & 90.52 & 91.28 & 94.09 & 92.77 & 94.93 & 88.00 & 91.10 & 87.27 & 90.22 & \textbf{94.27} & \textbf{96.55} \\
pill       & 43.28 & 47.11 & 46.99 & 49.76 & 41.52 & 43.54 & 47.32 & 58.31 & 87.19 & 90.05 & 83.21 & 89.00 & 84.69 & 90.90 & \textbf{89.02} & \textbf{92.99} \\
screw      & 25.10 & 31.35 & 46.96 & 59.03 & 35.94 & 42.37 & 47.12 & 55.17 & 46.65 & 50.79 & 38.47 & 49.49 & 37.32 & 48.07 & \textbf{53.36} & \textbf{64.27} \\
tile       & 85.33 & 91.60 & 89.21 & 93.74 & 85.67 & 90.28 & 83.53 & 87.30 & 88.87 & 91.96 & 84.29 & 89.72 & 82.42 & 89.25 & \textbf{90.32} & \textbf{94.77} \\
toothbrush & 39.40 & 63.93 & 65.35 & 79.43 & 53.75 & 60.46 & 57.68 & 72.03 & 61.00 & 70.50 & 48.68 & 64.41 & 35.36 & 40.98 & \textbf{72.77} & \textbf{91.58} \\
transistor & 65.03 & 71.05 & 59.96 & 62.18 & 29.28 & 30.67 & 63.71 & 66.79 & 73.56 & 78.48 & 79.27 & 91.74 & 77.96 & 81.75 & \textbf{89.60} & \textbf{94.24} \\
wood       & 49.64 & 60.47 & 67.52 & 73.28 & 50.91 & 53.16 & 61.84 & 89.54 & 67.00 & 80.84 & 60.16 & 74.62 & 56.90 & 62.33 & \textbf{78.85} & \textbf{88.34} \\
zipper     & 65.39 & 71.89 & 69.29 & 79.36 & 69.98 & 79.31 & 68.78 & 78.50 & 66.34 & 70.50 & 65.36 & 72.66 & 67.09 & 78.29 & \textbf{72.46} & \textbf{81.65} \\
\hline
Average    & 55.87 & 63.81 & 66.86 & 74.75 & 56.23 & 61.44 & 62.89 & 71.70 & 67.71 & 74.07 & 62.88 & 72.06 & 60.52 & 67.27 & \textbf{74.53} & \textbf{84.08} \\
\hline
\end{tabular}
}
\label{Segformer_mvtec}
\vskip -0.1in
\end{table*}

The generator is trained using a composite loss that includes global adversarial feedback, foreground-aware adversarial guidance, and a region-weighted reconstruction loss. The full objective is given by:
\begin{equation}
\mathcal{L}_G = \lambda_{\text{img}} \cdot \mathcal{L}_{\text{adv-img}} + \lambda_{\text{mask}} \cdot \mathcal{L}_{\text{adv-mask}} + \alpha \cdot \mathcal{L}_{\text{MSE}},
\end{equation}
where:
\begin{align}
\mathcal{L}_{\text{adv-img}} &= \mathbb{E}_{\mathbf{x} \sim p_g}\left[\phi\big(-D_{\text{img}}(\mathbf{x})\big)\right], \\
\mathcal{L}_{\text{adv-mask}} &= \mathbb{E}_{\mathbf{x} \sim p_g}\left[\phi\big(-D_{\text{fg}}(\mathbf{x}, \mathbf{m})\big)\right],
\end{align}
and the reconstruction loss is defined as:
\begin{equation}
\mathcal{L}_{\text{MSE}} = \left\|\mathbf{m} \odot (\hat{\mathbf{x}}_0 - \mathbf{x}_0)\right\|_2^2 + \beta \cdot \left\|(1 - \mathbf{m}) \odot (\hat{\mathbf{x}}_0 - \mathbf{x}_0)\right\|_2^2.
\end{equation}
Here, \( \hat{\mathbf{x}}_0 \) is the generator output, \( \mathbf{x}_0 \) is the original clean image, and \( \mathbf{m} \in \{0,1\}^{H \times W} \) is the anomaly mask. The softplus activation \( \phi(x) = \log(1 + e^x) \) is used for both adversarial terms. The coefficient \( \beta < 1 \) down-weights the background reconstruction penalty to encourage more accurate anomaly synthesis in foreground regions.

Unless otherwise specified, we use \( \lambda_{\text{img}} = 1.0 \), \( \lambda_{\text{mask}} = 1.0 \), \( \alpha = 1.0 \), and \( \beta = 0.1 \) in all experiments.

\section{Experiments and Results}

\subsection{Implementation Details}

\textbf{Datasets.} We conduct experiments on two widely-used industrial anomaly segmentation benchmarks: \textbf{MVTec-AD}~\cite{bergmann2019mvtec} and \textbf{BTAD}~\cite{btad2020}. For each anomaly category, we synthesize 500 image–mask pairs using only normal images and randomly generated binary masks. Approximately one-third of the generated samples are used to train the segmentation network, while the remaining two-thirds are held out for evaluation.

\textbf{Evaluation Metrics.} We evaluate segmentation performance using \textbf{mean Intersection over Union (mIoU)} and \textbf{pixel-wise accuracy (Acc)}, computed between predicted masks and ground-truth anomalies. All metrics are averaged across anomaly categories.

\textbf{Baselines.} We compare our method with six representative anomaly synthesis approaches: CutPaste~\cite{li2021cutpaste}, DRAEM~\cite{zavrtanik2021draem}, GLASS~\cite{glass}, DFMGAN~\cite{dfmgan}, AnomalyDiffusion~\cite{hu2024anomalydiffusion}, and RealNet~\cite{zhang2024realnet}. Additionally, we include \textbf{DDGAN}~\cite{ddgan2022} as an architectural baseline.

\textbf{Segmentation Backbones.} To assess generalization across architectures, we evaluate all methods using two lightweight real-time segmentation networks: \textbf{SegFormer}~\cite{xie2021segformer} and \textbf{BiseNet V2}~\cite{yu2021bisenet}. Both models are trained with identical settings on synthetic data generated by each method.

\textbf{Training Configuration.} The generator is trained with 4 discrete diffusion steps using a batch size of 4. We use Adam optimizers with learning rates of $1.6 \times 10^{-4}$ (generator) and $1.0 \times 10^{-4}$ (discriminator). An exponential moving average (EMA) with decay rate 0.999 is applied to stabilize generation. The discriminator is updated and regularized via R1 penalty. Training is performed for 20{,}000 iterations on 2 NVIDIA A800 GPUs.

\subsection{Comparison Studies}

\begin{table*}[htbp]
\centering
\vskip -0.1in
\caption{Comparison of pixel-level anomaly segmentation (mIoU/Acc) using the real-time BiseNet V2 trained on synthetic MVTec data produced from the proposed SARD and other existing AS methods.}
\resizebox{1\textwidth}{!}{
\begin{tabular}{l|cc|cc|cc|cc|cc|cc|cc|cc}
\hline
\textbf{Category} 
  & \multicolumn{2}{c|}{\textbf{CutPaste}} 
  & \multicolumn{2}{c|}{\textbf{DRAEM}} 
  & \multicolumn{2}{c|}{\textbf{GLASS}} 
  & \multicolumn{2}{c|}{\textbf{RealNet}} 
  & \multicolumn{2}{c|}{\textbf{DFMGAN}} 
  & \multicolumn{2}{c|}{\textbf{AnomalyDiffusion}} 
  & \multicolumn{2}{c|}{\textbf{DDGAN}} 
  & \multicolumn{2}{c}{\textbf{SARD}} \\ \hline

  & \textbf{mIoU} $\uparrow$ & \textbf{Acc} $\uparrow$ 
  & \textbf{mIoU} $\uparrow$ & \textbf{Acc} $\uparrow$ 
  & \textbf{mIoU} $\uparrow$ & \textbf{Acc} $\uparrow$ 
  & \textbf{mIoU} $\uparrow$ & \textbf{Acc} $\uparrow$ 
  & \textbf{mIoU} $\uparrow$ & \textbf{Acc} $\uparrow$ 
  & \textbf{mIoU} $\uparrow$ & \textbf{Acc} $\uparrow$ 
  & \textbf{mIoU} $\uparrow$ & \textbf{Acc} $\uparrow$ 
  & \textbf{mIoU} $\uparrow$ & \textbf{Acc} $\uparrow$ \\ 
\hline
bottle      & 71.77 & 78.57 & 75.13 & 79.17 & 57.81 & 60.79 & 72.16 & 75.55 & 64.28 & 71.31 & 75.28 & 85.11 & 61.25 & 65.97 & \textbf{81.70} & \textbf{90.79} \\
cable       & 46.00 & 57.08 & 53.88 & 60.96 & 16.63 & 16.65 & 51.22 & 62.32 & 57.09 & 63.25 & 60.55 & 74.96 & 48.03 & 56.45 & \textbf{70.24} & \textbf{79.06} \\
capsule     & 25.97 & 37.04 & 36.82 & 42.19 & 19.53 & 51.89 & 35.97 & 39.39 & 28.40 & 31.18 & 26.77 & 32.87 & 33.49 & 42.64 & \textbf{51.72} & \textbf{64.24} \\
carpet      & 58.98 & 72.22 & 68.42 & 77.21 & 64.77 & 73.93 & 8.98  & 9.01  & 62.13 & 67.98 & 58.18 & 64.69 & 61.59 & 73.52 & \textbf{70.13} & \textbf{82.69} \\
grid        & 24.68 & 44.17 & \textbf{42.81} & \textbf{63.34} & 6.50  & 6.91  & 10.61 & 11.47 & 10.17 & 15.23 & 18.98 & 24.30 & 2.94  & 2.83  & 38.53 & 55.13 \\
hazel\_nut  & 47.93 & 53.57 & 74.83 & 81.35 & 71.54 & 75.62 & 60.16 & 65.93 & 79.78 & 84.37 & 57.26 & 70.41 & 55.37 & 62.76 & \textbf{78.34} & \textbf{85.22} \\
leather     & 31.11 & 58.36 & 55.07 & 61.58 & 57.98 & 71.84 & 53.77 & 63.85 & 31.77 & 34.82 & 50.02 & 61.60 & 50.87 & 62.15 & \textbf{68.46} & \textbf{78.62} \\
metal\_nut  & 82.95 & 87.73 & 91.58 & 94.73 & 83.82 & 85.42 & 88.38 & 90.73 & 91.17 & 93.57 & 85.52 & 90.20 & 79.55 & 84.46 & \textbf{92.57} & \textbf{96.16} \\
pill        & 55.62 & 67.04 & 45.23 & 48.99 & 23.88 & 24.15 & 72.59 & 86.32 & 82.40 & 84.30 & 80.87 & 87.02 & 79.74 & 83.21 & \textbf{88.13} & \textbf{93.71} \\
screw       & 4.88  & 6.63  & 25.08 & 35.77 & 12.32 & 13.11 & 22.35 & 23.78 & \textbf{38.14} & 40.36 & 23.23 & 29.91 & 30.65 & 40.21 & 33.01 & \textbf{46.06} \\
tile        & 76.25 & 85.75 & 86.17 & 90.45 & 77.32 & 80.28 & 77.16 & 84.84 & 85.69 & 90.12 & 79.32 & 85.63 & 76.75 & 81.96 & \textbf{87.77} & \textbf{93.12} \\
toothbrush  & 35.69 & 50.45 & 57.66 & 79.15 & 38.86 & 51.97 & 32.38 & 37.88 & 48.83 & 58.76 & 44.33 & 69.32 & 30.25 & 36.81 & \textbf{70.77} & \textbf{90.63} \\
transistor  & 44.48 & 51.79 & 59.88 & 65.96 & 44.93 & 53.04 & 61.68 & 68.59 & 76.52 & 82.13 & 76.34 & 89.94 & 72.89 & 77.37 & \textbf{85.39} & \textbf{91.36} \\
wood        & 35.51 & 46.00 & 49.82 & 62.09 & 36.41 & 51.10 & 47.29 & 61.35 & 51.84 & 63.70 & 52.06 & 72.75 & 52.48 & 60.68 & \textbf{74.26} & \textbf{80.58} \\
zipper      & 51.61 & 63.09 & 66.88 & 75.75 & 61.99 & 70.07 & 66.09 & 77.54 & 60.61 & 71.11 & 57.86 & 67.64 & 59.43 & 71.57 & \textbf{69.28} & \textbf{79.89} \\
\hline
Average     & 46.23 & 57.30 & 59.28 & 67.91 & 44.95 & 52.45 & 50.72 & 57.24 & 57.92 & 63.48 & 56.44 & 67.09 & 53.02 &60.17 & \textbf{70.57} & \textbf{80.39} \\ 
\hline
\end{tabular}
}
\label{mvtec_bise}
\end{table*}

\begin{table*}[htbp]
  \centering
  \vskip -0.1in
  \caption{Evaluation of pixel-level segmentation accuracy on extended BTAD data using Segformer and BiseNet V2.}
  \vspace{-5pt}
  \label{tab:btad_clean}
  \resizebox{1\textwidth}{!}{
  \begin{tabular}{c|c|cc|cc|cc|cc|cc|cc|cc|cc}
    \hline
    Backbone & Category
      & \multicolumn{2}{c|}{CutPaste}
      & \multicolumn{2}{c|}{DRAEM}
      & \multicolumn{2}{c|}{GLASS}
      & \multicolumn{2}{c|}{DFMGAN}
      & \multicolumn{2}{c|}{RealNet}
      & \multicolumn{2}{c|}{AnomalyDiffusion}
      & \multicolumn{2}{c|}{DDGAN}
      & \multicolumn{2}{c}{SARD} \\
    \hline
     & & mIoU $\uparrow$ & Acc $\uparrow$
       & mIoU $\uparrow$ & Acc $\uparrow$
       & mIoU $\uparrow$ & Acc $\uparrow$
       & mIoU $\uparrow$ & Acc $\uparrow$
       & mIoU $\uparrow$ & Acc $\uparrow$
       & mIoU $\uparrow$ & Acc $\uparrow$
       & mIoU $\uparrow$ & Acc $\uparrow$
       & mIoU $\uparrow$ & Acc $\uparrow$ \\
    \hline
    \multirow{3}{*}{Segformer} 
     & 01 & 66.94 & 78.20 & 67.86 & 80.14 & 68.02 & 79.57 & 67.02 & 78.03 & 67.17 & 80.20 & 66.55 & 76.31 & 65.35 & 74.37 & \textbf{75.08} & \textbf{84.69} \\
     & 02 & 65.04 & 83.64 & 69.52 & 82.96 & 69.99 & 83.58 & 68.75 & 84.92 & 70.64 & 83.90 & 68.06 & 84.74 & 60.09 & 71.70 & \textbf{69.86} & \textbf{81.21} \\
     & 03 & 50.96 & 60.41 & 50.39 & 54.30 & 51.77 & 53.53 & 38.95 & 41.55 & 48.76 & 57.50 & 54.85 & 80.20 & 69.26 & 76.03 & \textbf{78.22} & \textbf{85.47} \\
    \multirow{3}{*}{BiseNet V2} 
     & 01 & 57.15 & 69.88 & 49.16 & 63.48 & 44.09 & 50.57 & 49.49 & 59.20 & 45.45 & 57.65 & 46.66 & 55.18 & 48.42 & 58.20 & \textbf{57.17} & \textbf{68.55} \\
     & 02 & 59.45 & 82.05 & 66.46 & 80.29 & 66.37 & 79.46 & 66.02 & 79.21 & 66.11 & 81.67 & 65.57 & 84.00 & 57.53 & 71.11 & \textbf{67.34} & \textbf{81.01} \\
     & 03 & 31.84 & 40.62 & 36.15 & 39.04 & 30.80 & 37.15 & 20.12 & 21.48 & 29.55 & 33.11 & 42.27 & 74.41 & 68.09 & 80.39 & \textbf{76.35} & \textbf{91.08} \\
    \hline
  \end{tabular}
}
\end{table*}
\textbf{Anomaly Segmentation.} To evaluate the effectiveness of SARD, we synthesize 500 image–mask pairs for each anomaly type on both MVTec-AD and BTAD using only normal images and binary masks. One-third of these synthetic samples are combined with real labeled pairs for training, and the remaining real samples are used for evaluation. We benchmark SARD against a comprehensive set of anomaly synthesis baselines. All methods are evaluated using SegFormer~\cite{xie2021segformer} and BiseNet V2~\cite{yu2021bisenet} as segmentation backbones to ensure consistent comparison.

The results on MVTec-AD are reported in Table~\ref{Segformer_mvtec}. SARD consistently outperforms all baselines in terms of average mIoU and pixel-wise accuracy across the 15 object categories. In particular, on challenging classes such as \textit{capsule}, SARD improves the segmentation mIoU by 9.86\% and 14.90\% over the second-best methods when using SegFormer and BiseNet V2, respectively. For highly textured or subtle anomaly types like \textit{grid} and \textit{tile}, SARD achieves noticeable gains in structural alignment and background consistency. These results highlight the benefits of our Region-Constrained Diffusion (RCD), which restricts generative updates to foreground regions, and Discriminative Mask Guidance (DMG), which enhances local texture fidelity through region-aware adversarial supervision.

We further validate the generalization of SARD on the BTAD dataset, as shown in Table~\ref{tab:btad_clean}. SARD again delivers superior performance across all categories under both segmentation backbones. Notably, on class \textit{03}, which contains fine-grained and spatially sparse defects, SARD achieves an mIoU of 78.22 with SegFormer and 76.35 with BiseNet V2, surpassing all baselines by a large margin. These improvements demonstrate the effectiveness of segmentation-aware Anomaly Synthesis in realistic industrial conditions.

\begin{figure}[hbt]
  \centering
  \includegraphics[width=\linewidth,pagebox=artbox]{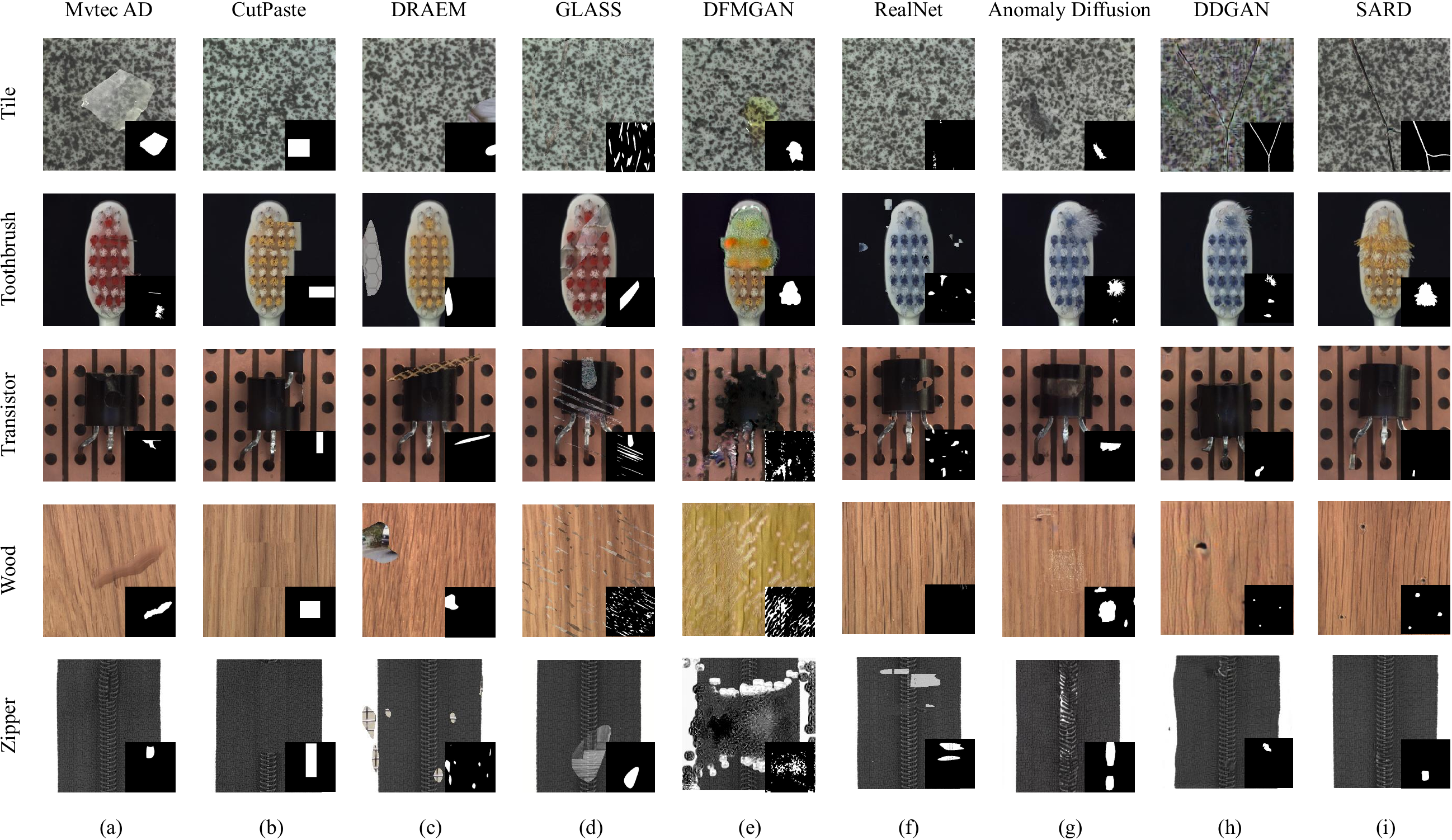}
  % 如果 artbox 不生效，可改为 pagebox=trimbox 或 pagebox=cropbox 试试
  \vspace{-16pt}
  \caption{%
    Visual comparison on MVTec-AD. Each column is a method
    (CutPaste, DRAEM, GLASS, RealNet, DFMGAN, AnomalyDiffusion, DDGAN, SARD).
    Results highlight boundary precision and texture quality.}
  \label{fig:qualitative_comparison}
  \vspace{-12pt}
\end{figure}

\textbf{Anomaly Synthesis Quality.} Fig.~\ref{fig:qualitative_comparison} compares visual results of different anomaly synthesis methods on representative MVTec categories. Traditional methods like CutPaste and DRAEM often generate artifacts with sharp transitions or unnatural boundaries. GAN-based approaches such as RealNet and DFMGAN improve texture realism but may introduce semantic misalignment or background inconsistencies.

In contrast, SARD produces anomaly textures that are mask-aligned, structurally coherent, and well integrated with the surrounding context. The synthesized anomalies exhibit high visual fidelity and accurate boundary adherence, especially in categories like \textit{metal\_nut}, \textit{tile}, and \textit{transistor}, demonstrating the advantages of RCD and DMG in achieving segmentation-oriented synthesis.

\subsection{Ablation Studies}

\textbf{Effect of Region-Constrained Diffusion (RCD).}
We first assess the impact of RCD by comparing against a baseline variant where the entire reverse denoising process follows the standard DDPM formulation without background preservation. That is, the full image is updated at each step, and no region-wise fusion is applied. As shown in Table~\ref{tab:ablation}, enabling RCD consistently improves mIoU and pixel-wise accuracy across most categories. For example, in \textit{bottle}, mIoU increases from \textbf{66.75} to \textbf{77.19}, and in \textit{capsule}, from \textbf{40.42} to \textbf{54.25}. These improvements demonstrate that restricting the denoising process to anomaly regions helps preserve background integrity and yields sharper, more localized defects.

\textbf{Effect of Discriminative Mask Guidance (DMG).}
We then examine the effect of DMG by removing the mask-guided foreground branch from the discriminator, leaving only global image-level supervision. Without region-specific feedback, the generator tends to produce less coherent structures. Adding DMG improves performance in categories with complex textures or irregular boundaries. For instance, in \textit{grid}, mIoU rises from \textbf{47.56} to \textbf{51.20}, and in \textit{leather}, from \textbf{58.51} to \textbf{67.45}. This highlights the effectiveness of localized adversarial learning in guiding anomaly synthesis towards finer boundary alignment and textural fidelity.

Therefore, RCD and DMG provide distinct but complementary advantages: RCD ensures structural consistency by freezing the background during generation, while DMG enhances foreground realism via spatially aware discrimination. Their combination: the complete SARD yields the strongest results across all categories, demonstrating the necessity of jointly modeling background preservation and region-specific supervision in diffusion-based anomaly synthesis.

\begin{table}[H]
\centering
\vskip -0.1in
\caption{Ablation study of RCD and DMG components in SARD using SegFormer on MVTec-AD.}
\label{tab:ablation}
\resizebox{0.5\textwidth}{!}{%
\begin{tabular}{l|cc|cc|cc|cc}
\toprule
 \textbf{Category} 
& \multicolumn{2}{c|}{\textbf{DDGAN (Baseline)}} 
& \multicolumn{2}{c|}{\textbf{w/ RCD}} 
& \multicolumn{2}{c|}{\textbf{w/ DMG}} 
& \multicolumn{2}{c}{\textbf{SARD}} \\
\midrule
& \textbf{mIoU} $\uparrow$ & \textbf{Acc} $\uparrow$ 
& \textbf{mIoU} $\uparrow$ & \textbf{Acc} $\uparrow$ 
& \textbf{mIoU} $\uparrow$ & \textbf{Acc} $\uparrow$ 
& \textbf{mIoU} $\uparrow$ & \textbf{Acc} $\uparrow$ \\
\midrule
bottle       & 66.75 & 70.70 & 77.19 & 80.22 & 80.61 & 84.29 & \textbf{83.52} & \textbf{87.71} \\
cable        & 55.83 & 61.67 & 64.12 & 63.41 & 62.60 & 68.22 & \textbf{70.76} & \textbf{76.60} \\
capsule      & 40.42 & 48.74 & 54.25 & 53.82 & 51.59 & 61.92 & \textbf{61.25} & \textbf{71.04} \\
carpet       & 67.98 & 79.25 & 71.04 & 74.19 & 73.31 & 83.88 & \textbf{76.39} & \textbf{84.77} \\
grid         & 47.56  & 47.99  & 51.80 & 61.45 & 51.20 & 57.88 & \textbf{53.15} & \textbf{67.56} \\
hazel\_nut   & 63.77 & 68.33 & 74.72 & 81.04 & 69.05 & 73.12 & \textbf{79.30} & \textbf{82.50} \\
leather      & 58.51 & 70.41 & 64.48 & 69.26 & 67.45 & 75.96 & \textbf{72.38} & \textbf{80.16} \\
metal\_nut   & 87.27 & 90.22 & 88.84 & 88.64 & 91.18 & 93.08 & \textbf{94.27} & \textbf{96.55} \\
pill         & 84.69 & 90.90 & 86.71 & 86.43 & 85.69 & 90.13 & \textbf{89.02} & \textbf{92.99} \\
screw        & 37.32 & 48.07 & 46.81 & 55.29 & 44.44 & 56.02 & \textbf{53.36} & \textbf{64.27} \\
tile         & 82.42 & 89.25 & 84.39 & 86.70 & 89.16 & 93.75 & \textbf{90.32} & \textbf{94.77} \\
toothbrush   & 35.36 & 40.98 & 64.88 & 74.36 & 67.45 & 83.83 & \textbf{72.77} & \textbf{91.58} \\
transistor   & 77.96 & 81.75 & 81.16 & 87.55 & 84.46 & 84.68 & \textbf{89.60} & \textbf{94.24} \\
wood         & 56.90 & 62.33 & 69.79 & 80.45 & 76.94 & 87.86 & \textbf{78.85} & \textbf{88.34} \\
zipper       & 67.09 & 78.29 & 68.38 & 74.26 & 68.47 & 75.63 & \textbf{72.46} & \textbf{81.65} \\
\bottomrule
\end{tabular}%
}
\end{table}
\label{ab}
\vspace{-10pt}

% \begin{wrapfigure}{r}{0.60\textwidth}
%   \centering
%   \vspace{-10pt}
%   \includegraphics[width=\linewidth]{imgs/med_vis.pdf}
%   \caption{Effect of MED. Foreground-aware supervision improves texture sharpness and structural alignment of generated anomalies.}
%   \label{fig:med_visual}
%   \vspace{-12pt}
% \end{wrapfigure}

\section{Conclusion}

We presented SARD, a segmentation-aware anomaly synthesis framework designed for industrial scenarios. By incorporating Region-Constrained Diffusion (RCD) and Discriminative Mask Guidance (DMG), SARD effectively addresses common issues in generative anomaly synthesis, such as spatial misalignment and background distortion. RCD constrains the reverse diffusion process to foreground regions while preserving the background from forward diffusion, ensuring more precise localization. DMG enhances the discriminator with mask-aware supervision, guiding the generator to produce sharper and more semantically aligned anomalies. Extensive experiments on MVTec-AD and BTAD datasets, using multiple segmentation backbones, demonstrate that SARD consistently outperforms state-of-the-art baselines. Ablation studies further confirm that both RCD and DMG provide complementary benefits. These findings highlight the importance of jointly enforcing structural constraints and region-specific adversarial learning in diffusion-based anomaly synthesis.

\bibliographystyle{IEEEtran}
\bibliography{ref}
\end{document}